\documentclass[runningheads]{llncs}
\usepackage{graphicx}
\usepackage{amsmath}
\usepackage{amssymb}
\usepackage{color}
\usepackage{booktabs}
\usepackage{multirow}
\usepackage{booktabs}
\usepackage{floatrow}
\usepackage{xcolor}
\usepackage{floatrow}
\usepackage{fontawesome}

\newcommand{\cxb}[1]{\textcolor{black}{#1}}
\newcommand{\yzr}[1]{\textcolor{black}{#1}}
\newcommand{\yzri}[1]{\textcolor{black}{#1}}

\begin{document}
\title{Adaptive Margin Global Classifier for Exemplar-Free Class-Incremental Learning}
\titlerunning{Adaptive Margin Global Classifier for EFCIL}
%
\author{Zhongren Yao \and Xiaobin Chang\textsuperscript{(\faEnvelopeO)}}
\authorrunning{Z. Yao, X. Chang}
%
\institute{School of Artificial Intelligence, Sun Yat-sen University, China
\email{yaozhr5@mail2.sysu.edu.cn, changxb3@mail.sysu.edu.cn}\\}
\maketitle              
\begin{abstract}
Exemplar-free class-incremental learning (EFCIL) presents a significant challenge as the old class samples are absent for new task learning.
Due to the severe imbalance between old and new class samples, the learned classifiers can be easily biased toward the new ones.
Moreover, continually updating the feature extractor under EFCIL can compromise the discriminative power of old class features, e.g., leading to less compact and more overlapping distributions across classes.
Existing methods mainly focus on handling biased classifier learning.
In this work, both cases are considered using the proposed method.
Specifically, we first introduce a Distribution-Based Global Classifier (\emph{DBGC}) to avoid bias factors in existing methods, such as data imbalance and sampling.
More importantly, the compromised distributions of old classes are simulated via a simple operation, variance enlarging (\emph{VE}).
Incorporating VE based on DBGC results in a novel classification loss for EFCIL. This loss is proven equivalent to an Adaptive Margin Softmax Cross Entropy (\emph{AMarX}).
The proposed method is thus called Adaptive Margin Global Classifier (\emph{AMGC}).
AMGC is simple yet effective. Extensive experiments show that
AMGC achieves superior image classification results on its own under a challenging EFCIL setting.

\keywords{class-incremental learning \and exemplar-free \and 
marginal loss.}

\end{abstract}
\section{Introduction}
\label{submission}
\begin{figure}[t]
\vskip 0.2in
\begin{center}
\centerline{\includegraphics[width=\columnwidth]{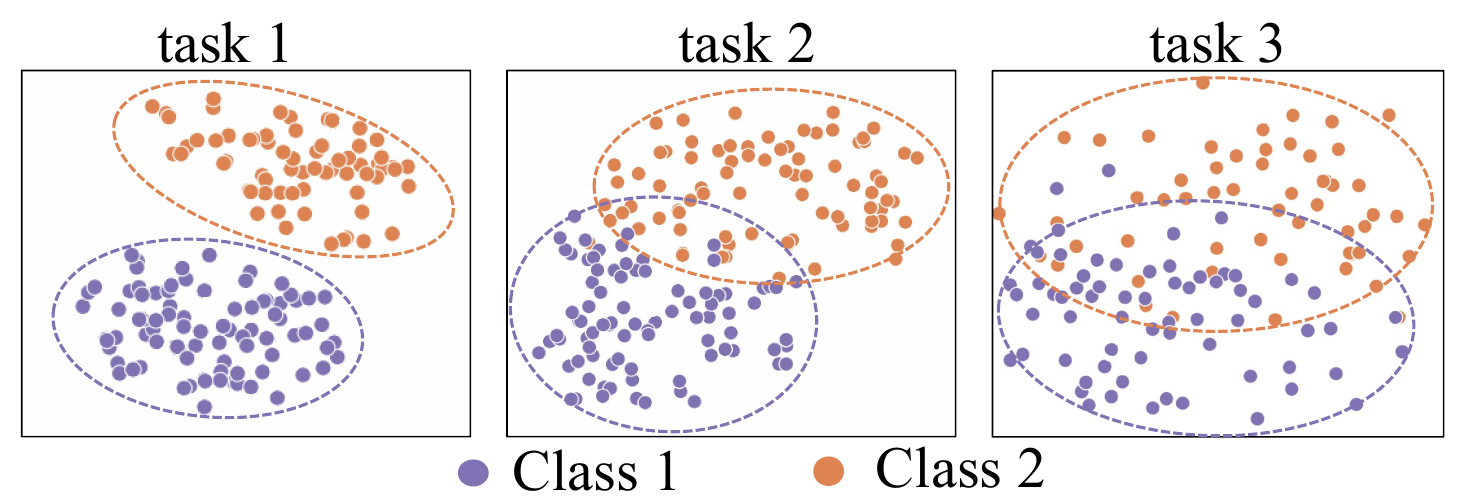}}
\caption{
Illustrations of the old class feature degradation
along with incremental learning.
The classification model learns Class 1 and Class 2 at task 1. Their features are compact and disjoint.
Under EFCIL, the model continually learned the new tasks, i.e., tasks 2 and 3, where Class 1 and Class 2 are old classes. Their features degrade to be more divergent and overlapped.
}
\label{img:feature}
\end{center}
\vskip -0.2in
\end{figure}

Class-incremental learning (CIL) is a challenging classification setting where training samples of novel classes are continually introduced within new tasks.
Under CIL, models are sequentially trained on new tasks and expected to accumulate knowledge, resulting in superior accuracy in both old and new classes.
However, severe performance degradation on the previously seen classes is observed, known as catastrophic forgetting~\cite{goodfellow2014empirical,mccloskey1989catastrophic}.
Due to user privacy or device limitations in practice, preserving and replaying exemplars from previous tasks as in the Exemplar-based methods~\cite{chen2023dynamic,jeeveswaran2023birt,rebuffi2017icarl} can be infeasible.
To this end, this paper focuses on a more challenging setting, Exemplar-free class-incremental learning (EFCIL)~\cite{petit2023fetril,zhu2021class}, where old class samples cannot be preserved.
EFCIL poses two main difficulties to classification algorithms.
Firstly, classifiers exclusively trained on new task samples tend to exhibit bias for new classes~\cite{petit2023fetril}.
Secondly, continual learning of the feature extractor in the EFCIL data stream can degrade the feature distributions of old classes~\cite{wu2021striking}, resulting in less compact and more overlapping feature distributions, as shown in Figure~\ref{img:feature}.

Various methods have been developed to mitigate the biased classifier learning issue in EFCIL.
One such approach aims to compensate for the absence of old class samples by generating pseudo features from the statistics (such as prototypes) of old classes~\cite{zhu2021prototype,zhu2022self,petit2023fetril}. These pseudo features and the extracted features of new class samples are used for more balanced global classifier training.
The learning of old and new classifiers can also be handled separately.
A naive solution could be freezing the old classifiers and training the new ones with new
task data.
The statistics of the old classes (i.e., prototype features and covariance matrices) further enable the training of the old classifiers during the new task~\cite{zhu2021class}.
Another approach abandons learning the classifier head and instead derives metric distances in the feature space to enable classification~\cite{goswami2023fecam}.

While training the feature extractor during the incremental process of EFCIL, old class features can suffer a severe loss of discriminative power and result in compromised distributions.
This is due to the extreme data imbalance in the new task where old class samples are completely absent~\cite{wu2021striking}.
However, existing EFCIL methods seem not to pay enough attention to this vital issue and
the reasons can be twofold.
On the one hand, the benchmark EFCIL settings assume either a large initial task 
, e.g., including data from half of all classes, is available~\cite{goswami2023fecam,petit2023fetril,zhu2021class,zhu2021prototype}
or a foundation model 
such as the vision transformer pretrained on ImageNet is based~\cite{mcdonnell2023ranpac,wang2023hierarchical,smith2023coda}.
The degradation of old class features can be alleviated with such strong feature extractors.
On the other hand,
methods with frozen feature extractors at initial states~\cite{goswami2023fecam,petit2023fetril} consistently outperform those with continually learned feature extractors~\cite{zhu2021class,zhu2021prototype,zhu2022self}.
It suggests that effective learning of the feature extractor remains a challenge under EFCIL.

In this paper, we propose a novel classification model that takes the aforementioned issues into consideration.
Specifically,
based on the statistics of seen (both old and new) classes, including mean prototypes and covariance matrices,
a Distribution-Based Global Classifier (\emph{DBGC}) is introduced.
DBGC mitigates the classifier biases from data sample imbalance, local optima~\cite{zhu2021class}, and pseudo feature sampling~\cite{petit2023fetril,zhu2021prototype,zhu2022self}.
Moreover, the proposed method considers the compromised feature distributions of old classes and simulates them with variance enlarging (\emph{VE}). VE simply enlarges the values of old class covariance matrix diagonals.
A novel classification loss for EFCIL has been proposed by integrating VE with DBGC.
We prove that this loss is equivalent to a softmax cross entropy with adaptive margins for old classes and refer to it as Adaptive Margin Softmax Cross Entropy (\emph{AMarX}).
AMarX also implies that when learning a classification model under EFCIL, one should be aware of the dynamics of the old class features and keep safe margins.
Our full model is thus called Adaptive Margin Global Classifier (\emph{AMGC}).
The main contributions are summarized as follows:
\begin{itemize}
\item The proposed AMGC is a simple yet effective classification model for EFCIL. It is built upon a Distribution-Based Global Classifier (DBGC) to mitigate the biases that arise from sampling and local optima.

\item The effect of degradation in old class features should be considered under EFCIL. We first simulate it through the variance enlarging (VE) operation and then seamlessly integrate VE into DBGC, resulting in a new classification loss called Adaptive Margin Softmax Cross Entropy (AMarX). AMarX has proven to be able to adjust the margins of the respective classes.

\item To highlight incremental learning procedures and reduce the impacts of strong initial models, experiments are mainly conducted under a challenging EFCIL setting.
The effectiveness of AMGC is demonstrated by the state-of-the-art (SOTA) performance and examined with detailed analysis.
\end{itemize}

\section{Related Work}
\cxb{Class Incremental Learning (CIL) is an important setting under continual learning~\cite{de2021continual,lin2023theory,van2019three}, which is a broader research topic.
The CIL methods aim to equip deep models with the capacity to learn from sequential tasks with disjoint classes and defy the catastrophic forgetting problem~\cite{goodfellow2014empirical,mccloskey1989catastrophic}.
To maintain the knowledge of previous tasks, the exemplar-based CIL (EBCIL)~\cite{chen2023dynamic,jeeveswaran2023birt,rebuffi2017icarl} allows preserving limited training samples of previous tasks as exemplars and replaying them at new task learning.}

\subsection{Exemplar-Free Class Incremental Learning}


\cxb{In exemplar-free CIL (EFCIL)~\cite{petit2023fetril,zhu2021class}, no exemplar is preserved and replayed at new task learning.}
To alleviate the catastrophic forgetting in feature learning, a regularization based on posterior estimations~\cite{yan2021dynamically} controls crucial changes in model parameters.
An assumption that the parameter changes across tasks should be restricted in the local region is applied in EWC~\cite{kirkpatrick2017overcoming}.
Knowledge Distillation~\cite{hinton2015distilling} can be used to transfer knowledge from previous models to the current one at the new task learning~\cite{zhu2022self}.
With the foundation model available, prompt-based methods
\yzr{~\cite{smith2023coda,gao2024consistent}}
are proposed for efficient adaptation and transfer.
Recent studies~\cite{goswami2023fecam,petit2023fetril} have shown that state-of-the-art results can be obtained by freezing the feature extractors which were well-pretrained on a large initial task.
\cxb{In this work, we conduct experiments under a more challenging EFCIL setting with much smaller initial data and the model training from scratch.}
The absence of old class samples in EFCIL \cxb{also} poses a significant challenge in learning an unbiased classifier head.
Instead of learning a parameterized classifier, a distance metric based on covariance matrices is proposed~\cite{goswami2023fecam}.
Another direct solution can be generating pseudo features of old classes as compensation. For example, such augmented features can be sampled based on old class statistics~\cite{zhu2021prototype,zhu2022self} or transferred from new classes~\cite{petit2023fetril}.
To avoid the sampling bias introduced by the feature generation, a distribution-based loss~\cite{wang2021regularizing} for supervised learning is adopted by IL2A~\cite{zhu2021class} to handle the learning of old classifiers at new tasks. However, the old and new classifiers are learned separately in IL2A and can be limited by the local optima.
The proposed Distribution-Based Global Classifier (DBGC) unifies the learning of old and new classifiers under a distribution-based loss. This approach achieves superior performance by learning a less biased holistic classifier.
More importantly, DBGC can be further advanced to a novel
loss, called Adaptive Margin Softmax Cross Entropy (AMarX), by considering the old class feature degradation.

\subsection{Classification Loss with Margin}

Introducing margin into a classification loss aims to enhance the separations between different categories \yzr{\cite{wan2018rethinking}}.
The frequently-used losses that integrated with margins, e.g., softmax cross-entropy~\cite{liang2017soft,liu2016large} and $k$-nearest neighbour~\cite{weinberger2009distance}, are found effective in applications such as face recognition~\cite{deng2019arcface,schroff2015facenet,wang2018cosface}.
The continual learning of classification tasks is also investigated as solving a sequential max-margin problem~\cite{evron2023continual}.
However, the proposed AMarX differs from the existing losses in two perspectives.
On the one hand, AMarX derives from reminding the model training of the old class feature degradation via simulating it. It thus serves very different purposes to its counterparts.
On the other hand, the margins of AMarX are adaptive to specific classes while existing ones are fixed for all classes.

\section{Adaptive Margin Global Classifier}

\begin{figure*}[t]
\vskip 0.2in
\begin{center}
\centering
\centerline{\includegraphics[width=\textwidth]{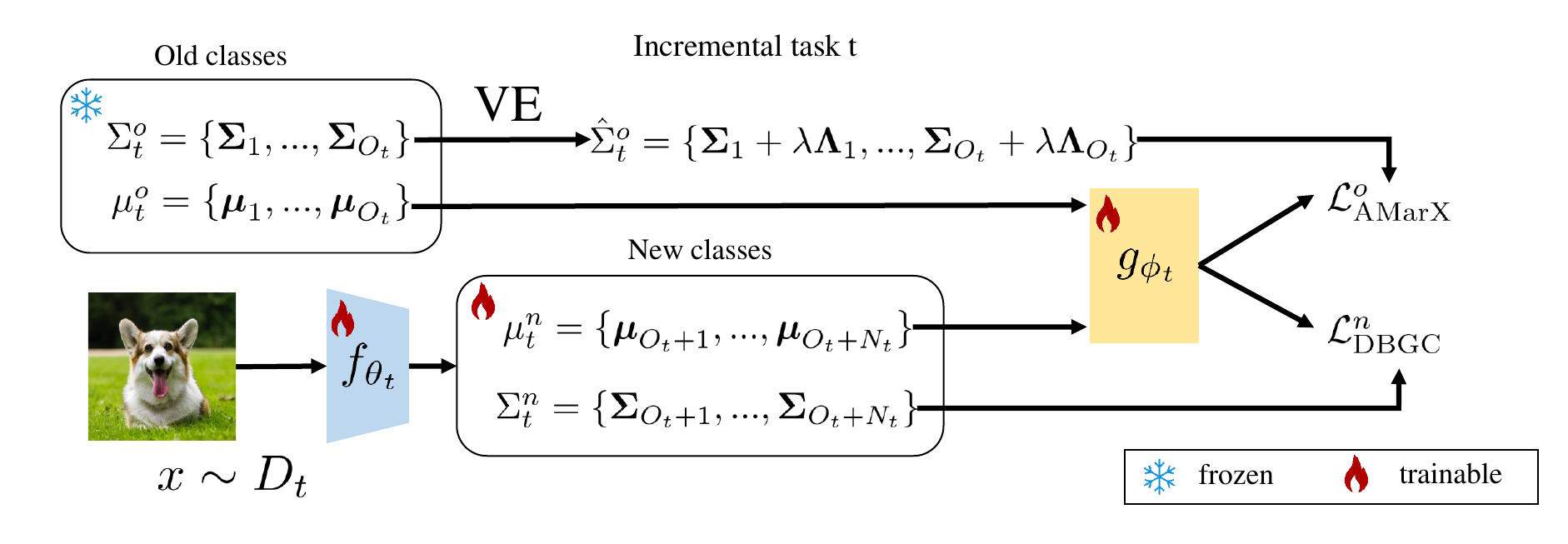}}
\caption{
\cxb{Illustration of the AMGC components.}
The Distribution-Based (DB) classification loss is derived and enables the learning of a global classifier (GC) entirely based on the statistics ($\mu_t$ and $\Sigma_t$) of both old and new classes,
as detailed in Section~\ref{sec:DBGC}.
Secondly,
DBGC incorporates the $\hat{\Sigma}_t^{o}$ from variance enlarging (VE), resulting in the new loss, AMarX,
for the old classes, as described in Section~\ref{sec:AMarX}.
}
\label{fig:framework}
\end{center}
\vskip -0.2in
\end{figure*}

The proposed Adaptive Margin Global Classifier (AMGC) consists of two parts.
Firstly, to handle the bias factors of classifier learning in EFCIL, a Distribution-Based Global Classifier (DBGC) is introduced, as described in Section~\ref{sec:DBGC}.
Secondly, the variance enlarging (VE) technique is exploited to simulate the degradation of old class features.
By integrating VE into DBGC, a novel classification loss called Adaptive Margin Softmax Cross Entropy (AMarX) is obtained, as detailed in Section~\ref{sec:AMarX}.
\cxb{The full model is depicted in Figure~\ref{fig:framework}.}
Moreover, necessary backgrounds and notations are first presented in Section~\ref{sec:pre-def}. 

\subsection{Preliminaries}\label{sec:pre-def}

In the class incremental learning (CIL) setting, a classification model is trained on $T$ tasks sequentially.
Training data for task $t \in \{1,...,T\}$ is denoted by $D_{t}$ and covers the classes in $C_t$. There are $N_t$ different classes in the class set $C_t$.
CIL requires $C_i \cap C_j = \emptyset$, $i \neq j, \forall i,j \in \{1,...,T\}$.
Within task $t$, \emph{new} classes are those from $C_t$ while \emph{old} classes are those from all previous tasks $\cup_{j=1}^{t-1} C_j$.
There are $N_t$ new classes and $O_t = \sum_{j=1}^{t-1} N_j$ old classes.
As the task identity is not available during CIL testing, a holistic label space along the incremental procedure is required. A straightforward solution is assigning each new class in $C_t$ to a unique label in $Y_t^n = \{O_t+1, ..., O_t+N_t\}$. The labels of the old classes naturally become $Y_t^o = \{1, ..., O_t\}$ at task $t$. The label space of \emph{seen} (both old and new) classes at task $t$ is $Y_t^s = Y_t^n \cup Y_t^o = \{1, ..., O_t, O_t+1, ..., O_t+N_t\}$.
The exemplar-free class incremental learning (EFCIL) can be a more challenging setting than CIL. Under EFCIL, the training samples of task $t$ are from $D_t$ only, while CIL allows a memory buffer to store the samples from previous tasks and replaying them at new tasks.
This paper follows a challenging EFCIL setting with tasks evenly split.
Specifically, the size of $D_t$ (or $C_t$) across different $t\in\{1,...,T\}$ are the same.

A classification model consists the feature extractor $f_{\theta}$ parameterized with $\theta$ and the classifier head $g_{\phi}$ parameterized with $\phi = (\textbf{W}, \textbf{b})$, $\textbf{W}$ indicates the classifier weights and $\textbf{b}$ is bias terms.
At the incremental task $t \in \{2,...,T\}$ of EFCIL, the classification model $\{f_{\theta_t}, g_{\phi_t}\}$ is trained with samples $(x,y) \sim D_t$, where $x$ indicates a raw input and the corresponding class label $y \in Y_t^n$. The feature of $x$ is $\textbf{\textit{f}} = f_{\theta_t}(x) \in R^d$.
$\theta_t$ is initialized with $\theta_{t-1}$ before training.
To predict the seen classes till $t$,
the shape of parameter in $\phi_{t} = (\textbf{W}_t, \textbf{b}_t)$ are thus $\textbf{W}_t \in R^{d \times (O_t+N_t)}$ and $\textbf{b}_t \in R^{O_t+N_t}$. To be more specific, $\textbf{W}_t = [\boldsymbol{\omega}_1, ..., \boldsymbol{\omega}_{O_t}, \boldsymbol{\omega}_{O_t+1}, ..., \boldsymbol{\omega}_{O_t+N_t}]= [\textbf{W}_t^o, \textbf{W}_t^n]$, where $\boldsymbol{\omega}_k \in R^d$, $k \in Y^s_t$ is the weight vector of class $k$. $\textbf{W}_t^o \in R^{d \times {O_t}}$ and $\textbf{W}_t^n \in R^{d \times {N_t}}$ are weights for the old and new classes respectively. Similarly, $\textbf{b}_t = [b_1; ...; b_{O_t}; b_{O_t+1}; ...; b_{O_t+N_t}]= [\textbf{b}_t^o; \textbf{b}_t^n]$ with $\textbf{b}_t^o \in R^{O_t}$ and $\textbf{b}_t^n \in R^{N_t}$.
The parameters of the old classes are $\phi_{t}^o = (\textbf{W}_t^o, \textbf{b}_t^o)$ and those of the new class are $\phi_{t}^n = (\textbf{W}_t^n, \textbf{b}_t^n)$.
Therefore, $\phi_t$ is either partially ($\phi_{t}^o$) initialized with $\phi_{t-1}$ or totally initialized from scratch.
At the initial task $t=1$, the model $f_{\theta_1}$, $g_{\phi_1}$ and their training simply follow the conventional classification pipeline. 

The statistics of class $k \in Y^s_t$ in the feature space, i.e., the mean vector $\boldsymbol{\mu}_k$ and the covariance matrix $\boldsymbol{\Sigma}_k$, can be exploited by the EFCIL methods.
Old class statistics from previous tasks are calculated with the corresponding trained feature extractor and training samples and are saved for future tasks.
New class statistics can be iteratively calculated along with the feature extractor training based on mini-batch samples\footnote{\cxb{Details of the online update are available in supplementary material section A.}}.
At task $t$, the statistics of old classes are denoted as $\mu_t^o=\{\boldsymbol{\mu}_1,...,\boldsymbol{\mu}_{O_t}\}$ and $\Sigma_t^o=\{\boldsymbol{\Sigma}_1,...,\boldsymbol{\Sigma}_{O_t}\}$.
The new class ones are $\mu_t^n=\{\boldsymbol{\mu}_{O_t+1},...,\boldsymbol{\mu}_{O_t+N_t}\}$ and $\Sigma_t^n=\{\boldsymbol{\Sigma}_{O_t+1},...,\boldsymbol{\Sigma}_{O_t+N_t}\}$.
The pseudo feature $\boldsymbol{\tilde{f}}_k$ of a class $k$ can be generated based on the statistics $\boldsymbol{\mu}_k$ and $\boldsymbol{\Sigma}_k$, i.e.,
sampling from a Gaussian prior $\boldsymbol{\tilde{f}}_k \sim \mathcal{N}(\boldsymbol{\mu}_k, \boldsymbol{\Sigma}_k)$ in this work.

\subsection{Distribution-Based Global Classifier}\label{sec:DBGC}

The Distribution-Based (DB) classification loss $\mathcal{L}_{\operatorname{DB}}$ is first introduced under a simplified scenario.
Assuming a classification problem with $K$ classes,
their statistics, mean vectors $\mu = \{\boldsymbol{\mu}_1,...,\boldsymbol{\mu}_K\}$ and covariance matrices $\Sigma = \{\boldsymbol{\Sigma}_1, ..., \boldsymbol{\Sigma}_K\}$ are available.
The parameters of classifier $g$ are $\phi = (\textbf{W}, \textbf{b})$, where $\textbf{W}\in R^{d \times K}$ and $\textbf{b}\in R^{K}$.
Based on the $M$ pseudo features of class $k$ sampled,
$\boldsymbol{\tilde{f}}_k \sim \mathcal{N}(\boldsymbol{\mu}_k, \boldsymbol{\Sigma}_k)$, the Sample-Based (SB) loss $\mathcal{L}_{\operatorname{SB}}^{M}$ is a softmax cross-entropy
\begin{equation}
\begin{split}
\label{eq:sample}
    \mathcal{L}_{\operatorname{SB}}^{M}(\mu,\Sigma; \theta, \phi) 
    &=\frac{1}{KM}\sum_{k=1}^{K}\sum_{i=1}^M\log(\sum_{j=1}^{K} e^{(\boldsymbol{\omega}_j-\boldsymbol{\omega}_k)^T {\boldsymbol{\tilde{f}}}_{k,i}+(b_j-b_k)})\\
    &=\frac{1}{K}\sum_{k=1}^{K}\frac{1}{M}\sum_{i=1}^M\log(\sum_{j=1}^{K} e^{\boldsymbol{v}_{j,k}^T {\boldsymbol{\tilde{f}}}_{k,i}+\delta_{j,k}}),
\end{split}
\end{equation}
where $\boldsymbol{v}_{j,k}=\boldsymbol{\omega}_j-\boldsymbol{\omega}_k$ and $\delta_{j,k}=b_j-b_k$.
When $M \rightarrow \infty$,
\begin{align}
    \mathcal{L}_{\operatorname{SB}}^{\infty}&=\frac{1}{K}\sum_{k=1}^{K} \mathbb{E}_{\boldsymbol{\tilde{f}}_{k}} (\log(\sum_{j=1}^{K} e^{\boldsymbol{v}_{j,k}^T {\boldsymbol{\tilde{f}}}_{k}+\delta_{j,k}})) \label{eq:lsb_m_inf} \\
    & \leq \frac{1}{K}\sum_{k=1}^{K} \log(\mathbb{E}_{\boldsymbol{\tilde{f}}_{k}} (\sum_{j=1}^{K} e^{\boldsymbol{v}_{j,k}^T {\boldsymbol{\tilde{f}}}_{k}+\delta_{j,k}})),  \label{eq:lsb_ub}
\end{align}
where the Jensen’s inequality is applied from Eq.~(\ref{eq:lsb_m_inf}) to Eq.~(\ref{eq:lsb_ub}). 
With the Moment generating function of Gaussian,
\begin{equation}
    \label{eq:moment-gen}
    \mathbb{E}_{\boldsymbol{\tilde{f}}_{k}} (e^{\boldsymbol{v}_{j,k}^T{\boldsymbol{\tilde{f}}}_{k}}) = e^{\boldsymbol{v}_{j,k}^T \boldsymbol{\mu}_k+ \frac{\boldsymbol{v}_{j,k}^T \boldsymbol{\Sigma}_k \boldsymbol{v}_{j,k}}{2}} , \boldsymbol{\tilde{f}}_{k} \sim \mathcal{N}(\boldsymbol{\mu}_k, \boldsymbol{\Sigma}_k),
\end{equation}
Eq.~(\ref{eq:lsb_ub}) can be rewrite as
\begin{equation}
\begin{split}
    \frac{1}{K}\sum_{k=1}^{K} \log(\sum_{j=1}^{K} e^{\boldsymbol{v}_{j,k}^T \boldsymbol{\mu}_k+ \frac{\boldsymbol{v}_{j,k}^T \boldsymbol{\Sigma}_k \boldsymbol{v}_{j,k}}{2}  + \delta_{j,k}})
    \triangleq \mathcal{L}_{\operatorname{DB}}(\mu,\Sigma; \theta, \phi).
    \label{eq:L-DB}
\end{split}
\end{equation}
The resulting loss in Eq.~(\ref{eq:L-DB}) is calculated based on the class statistics ($\mu$ and $\Sigma$) only and requires no sample, thus called distribution-based (DB) loss $\mathcal{L}_{\operatorname{DB}}$.

At the incremental task $t$, the statistics of both old and new classes are available. 
The corresponding DB losses are
\begin{equation}
\begin{split}
     \mathcal{L}_{\operatorname{DBGC}}^n &= \mathcal{L}_{\operatorname{DB}}(\mu_t^n,\Sigma_t^n; \theta_t, \phi_t)\\
    & = \frac{1}{N_t} \underbrace{\sum_{k=O_t+1}^{O_t+N_t}}_{new} \log(\underbrace{\textcolor{black}{\sum_{j=1}^{O_t+N_t}}}_{seen} e^{\boldsymbol{v}_{j,k}^T \boldsymbol{\mu}_k+ \frac{\boldsymbol{v}_{j,k}^T \boldsymbol{\Sigma}_k \boldsymbol{v}_{j,k}}{2}  + \delta_{j,k}}),
    \label{eq:L-DBGC-N}
\end{split}
\end{equation}
\begin{equation}
\begin{split}
    \mathcal{L}_{\operatorname{DBGC}}^o & = \mathcal{L}_{\operatorname{DB}}(\mu_t^o,\Sigma_t^o; \theta_t, \phi_t)\\
    & = \frac{1}{O_t} \underbrace{\sum_{k=1}^{O_t}}_{old} \log(\underbrace{\textcolor{black}{\sum_{j=1}^{O_t+N_t}}}_{seen} e^{\boldsymbol{v}_{j,k}^T \boldsymbol{\mu}_k+ \frac{\boldsymbol{v}_{j,k}^T \boldsymbol{\Sigma}_k \boldsymbol{v}_{j,k}}{2}  + \delta_{j,k}}).
    \label{eq:L-DBGC-O}
\end{split}
\end{equation}
The proposed losses offer two benefits for CIL.
On the one hand, learning based on the $\mathcal{L}_{\operatorname{DB}}$ loss alleviates both the data imbalance across classes and the sampling bias of features and instances.
On the other hand, both losses in Eq.~(\ref{eq:L-DBGC-N}) and Eq.~(\ref{eq:L-DBGC-O}) aim to holistically optimize $\phi_t$, the parameters of a global classifier (GC), rather than separately optimizing $\phi^n_t$ and $\phi^o_t$ of local classifiers (LC) respectively.
Therefore, the overall loss for the Distribution-Based Glocal Classifier (DBGC) can be more straightforward
\begin{equation}
\begin{split}
    \mathcal{L}_{\operatorname{DBGC}} & = \mathcal{L}_{\operatorname{DB}}(\mu_t^o\cup \mu_t^n,\Sigma_t^o\cup\Sigma_t^n; \theta_t, \phi_t)\\
    & = \frac{1}{O_t+N_t} \underbrace{\sum_{k=1}^{O_t+N_t}}_{seen} \log(\underbrace{\textcolor{black}{\sum_{j=1}^{O_t+N_t}}}_{seen} e^{\boldsymbol{v}_{j,k}^T \boldsymbol{\mu}_k+ \frac{\boldsymbol{v}_{j,k}^T \boldsymbol{\Sigma}_k \boldsymbol{v}_{j,k}}{2}  + \delta_{j,k}}).
\end{split}
\end{equation}

Based on the terms DB, SB, GC, and LC defined in this section, a few variants\cxb{\footnote{More details can be found in supplementary material section B.}} other than our DBGC can be used for EFCIL. They will be compared in the experiment.

\subsection{Adaptive Margin Softmax Cross Entropy}\label{sec:AMarX}

The classification model learned under the EFCIL setting is vulnerable to catastrophic forgetting due to the absence of training samples from the old classes. 
\cxb{As shown in Figure~\ref{img:feature},}
the features of the old classes become less discriminative at new tasks because their distributions become more divergent \cxb{after learning on new tasks}.
In this work, such feature dynamics of the old classes are simulated by enlarging their variances and achieved via the Variance Enlarge (VE) operation
\begin{equation}
\label{eq:ve}
\boldsymbol{\hat{\Sigma}}_k = \boldsymbol{\Sigma}_k + \lambda \boldsymbol{\Lambda}_k,
\end{equation}
where $\boldsymbol{\Sigma}_k$ is the covariance matrix of an old class $k \in Y_t^o$. $\boldsymbol{\Lambda}_k$ is the diagonal matrix of $\boldsymbol{\Sigma}_k$ and records the variance of each feature dimension. By simply setting $\lambda > 0$, a new statistic $\boldsymbol{\hat{\Sigma}}_k$ with enlarged variances is obtained.
Applying VE to all matrices in $\Sigma_t^o$, 
we have
\begin{equation}
\begin{split}
    \hat{\Sigma}_t^o & =\{\boldsymbol{\hat{\Sigma}}_1,...,\boldsymbol{\hat{\Sigma}}_{O_t}\} \\
    & = \{\boldsymbol{\Sigma}_1+\lambda\boldsymbol{\Lambda}_1,...,\boldsymbol{\Sigma}_{O_t}+\lambda\boldsymbol{\Lambda}_{O_t}\},
\end{split}
\end{equation}
where a single $\lambda$ is used for different classes.

Replacing the $\Sigma_t^o$ in $\mathcal{L}_{\operatorname{DBGC}}^{o}$ (Eq.~(\ref{eq:L-DBGC-O})) with $\hat{\Sigma}_t^o$ and results in
\begin{equation}
\begin{split}
    & \mathcal{L}_{\operatorname{DB}}(\mu_t^o, \textcolor{red}{\hat{\Sigma}_t^o}; \theta_t, \phi_t)\\
     = & \frac{1}{O_t} {\sum_{k=1}^{O_t}} \log({\textcolor{black}{\sum_{j=1}^{O_t+N_t}}} e^{\boldsymbol{v}_{j,k}^T \boldsymbol{\mu}_k+ \frac{\boldsymbol{v}_{j,k}^T \textcolor{red}{( \boldsymbol{\Sigma}_k +\lambda\boldsymbol{\Lambda}_{k} )} \boldsymbol{v}_{j,k}}{2}  + \delta_{j,k}}).
    \label{eq:L-DBGC-O-VE}
\end{split}
\end{equation}
It shows that VE and DBGC can be seamlessly integrated.

To enable further analysis, we rewrite the softmax cross entropy of class $k$
in Eq.~(\ref{eq:L-DBGC-O-VE}) with $\boldsymbol{v}_{j,k}=\boldsymbol{\omega}_j-\boldsymbol{\omega}_k$ and $\delta_{j,k}=b_j-b_k$ as follows
\begin{align}
& -\log\frac{e^{\boldsymbol{\omega}_{k}^T
\boldsymbol{\mu}_k+b_{k}}}{\sum\limits_{j=1}^{O_t+N_t} e^{\boldsymbol{\omega}_{j}^T \boldsymbol{\mu}_k+b_{j}+ \frac{\boldsymbol{v}_{j,k}^T ( \boldsymbol{\Sigma}_k + \textcolor{red}{\lambda\boldsymbol{\Lambda}_{k}} ) \boldsymbol{v}_{j,k}}{2}
}}\label{eq:margin-start}\\
= & -\log\frac{e^{\boldsymbol{\omega}_{k}^T
\boldsymbol{\mu}_k+b_{k} - \textcolor{red}{m_{k}}}}{e^{\boldsymbol{\omega}_{k}^T
\boldsymbol{\mu}_k+b_{k} - \textcolor{red}{m_{k}}} + \sum\limits_{j \neq k}^{O_t+N_t} e^{\boldsymbol{\omega}_{j}^T \boldsymbol{\mu}_k+b_{j}+ \sigma_{j,k} + \beta_{j,k}
}},\label{eq:margin}
\end{align}
where\footnote{\cxb{Detailed derivation from Eq.~(\ref{eq:margin-start}) to Eq.~(\ref{eq:margin}) can be found in supplementary material section C.}}
$m_k=\frac{\lambda}{2} \boldsymbol{\omega}_k^{T}\boldsymbol{\Lambda}_k\boldsymbol{\omega}_k$, 
$\sigma_{j,k} = \frac{\boldsymbol{v}_{j,k}^T \boldsymbol{\Sigma}_k  \boldsymbol{v}_{j,k}}{2}$,
and $\beta_{j,k}=\frac{\lambda}{2}(\boldsymbol{\omega}_j^{T}\boldsymbol{\Lambda}_k\boldsymbol{\omega}_j-\boldsymbol{\omega}_j^{T}\boldsymbol{\Lambda}_k\boldsymbol{\omega}_k-\boldsymbol{\omega}_k^{T}\boldsymbol{\Lambda}_k\boldsymbol{\omega}_j)$.
$\sigma_{j,k}$ and $\beta_{j,k}$ encode the high-order information.
Since $\boldsymbol{\Lambda}_k$ is a diagonal matrix that records the variances of class $k$ features, it is positive definite.
$m_k > 0$ is thus a margin \emph{adaptive to a specific class} $k$.
The proposed Adaptive Margin Softmax Cross Entropy (AMarX) becomes
\begin{equation}
\begin{split}
    \mathcal{L}_{\operatorname{AMarX}}^o & = \mathcal{L}_{\operatorname{DB}}(\mu_t^o, \textcolor{black}{\hat{\Sigma}_t^o}; \theta_t, \phi_t)\\
     &=  \frac{-1}{O_t} {\sum_{k=1}^{O_t}} \log\frac{e^{\boldsymbol{\omega}_{k}^T
\boldsymbol{\mu}_k+b_{k} - \textcolor{black}{m_{k}}}}{e^{\boldsymbol{\omega}_{k}^T
\boldsymbol{\mu}_k+b_{k} - \textcolor{black}{m_{k}}} + \sum\limits_{j \neq k}^{O_t+N_t} e^{\boldsymbol{\omega}_{j}^T \boldsymbol{\mu}_k+b_{j}+ \sigma_{j,k} + \beta_{j,k}
}}.
    \label{eq:L-AMarX-O}
\end{split}
\end{equation}

The proposed method, Adaptive Margin Global Classifier (AMGC), combines DBGC and AMarX, as illustrated in Figure~\ref{fig:framework}. Specifically, DBGC aims to tackle the classification biases under EFCIL. VE is proposed to simulate compromised distributions of old classes, resulting in a novel loss AMarX based on DBGC.
The overall loss is
\begin{equation}
\label{eq:separate}
    \mathcal{L}_{\operatorname{AMGC}}=\mathcal{L}^{n}_{\operatorname{DBGC}}+{\mathcal{L}^{o}_{\operatorname{AMarX}}},
\end{equation}
where $\mathcal{L}^{n}_{\operatorname{DBGC}}$ (Eq.~(\ref{eq:L-DBGC-N})) is based on the statistics of new classes and $\mathcal{L}^{o}_{\operatorname{AMarX}}$ (Eq.~(\ref{eq:L-AMarX-O})) is based on those of old classes. Both losses are used to optimize the global classifier head $g_{\phi_t}$ and the feature extractor $f_{\theta_t}$ under the objective
\begin{equation}
    \label{eq:obj}
    \min_{\phi_t, \theta_t} \mathcal{L}_{\operatorname{AMGC}}.
\end{equation}
Our method is 
learned with $\mathcal{L}_{\operatorname{AMGC}}$ only.

\section{Experiment}\label{sec:main-exp}

\subsection{Experimental Details}\label{sec:main-exp-detail}

\textbf{Datasets and Protocols.}
Experiments are conducted on three image classification benchmark datasets.
(1) ImageNet Subset~\cite{deng2009imagenet} \cxb{(denoted as ImageNet-S)} is a large-scale dataset. It contains 100 classes from the full ImageNet dataset~\cite{russakovsky2015imagenet}. Each class with 1,300 training images and 50 testing images.
(2) TinyImageNet~\cite{le2015tiny} is also a subset of ImageNet with 200 classes. Its images are in 64$\times$64 resolution. There are 500 and 50 images per class for training and testing, respectively.
(3) CIFAR100~\cite{krizhevsky2009learning} consists of 
100 classes, 32$\times$32 resolution images with 500 and 100 images per class for training and testing.

\yzr{
CIFAR100 and \yzr{ImageNet-S} have 100 classes, and their three incremental scenarios are: (1) T = 10 with 10 new classes per task; (2) T = 20 with 5 new classes per task.
Tiny-ImageNet has 200 classes. Its two incremental scenarios (T = 10, 20) are similarly set.
We do not have access to any pre-trained models or privileged data.}


\textbf{Evaluation Metric.}
\yzr{\cxb{Following~\cite{zhao2020maintaining,chen2023dynamic}}, \cxb{two} CIL metrics, the accuracy \cxb{of seen classes at} the last incremental task \cxb{(denoted as LA)}
and the average incremental accuracy \cxb{(denoted as AIA)},
are adopted to measure the model performance.}
The proposed method is evaluated on 
\yzri{3}
different runs and reports the averaged results.

\textbf{Implementation Details.}
Following~\cite{goswami2023fecam,petit2023fetril,zhu2022self}, We use ResNet-18~\cite{he2016deep} as the backbone network for all experiments.
Our implementation is based on PyCIL~\cite{zhou2023pycil}.
\textcolor{black}{
The proposed model is optimized using the same strategy on different datasets and settings.
The model is trained from scratch at the initial task with a learning rate starting at 1e-2 for 400 epochs.
At the training of incremental tasks, both the feature extractor (with batch normalization layers fixed at the initial states) and the classifier head are continually optimized with lower learning rates (1e-6 and 5e-3 respectively) and fewer epochs (200 epochs).
}
\yzri{The proposed AMGC}
is concise, with $\lambda$ as the main hyper-parameter. We set $\lambda=0.4$.

\textbf{Competitors.}
Our AMGC is compared with the representative and state-of-the-art (SOTA) EFCIL methods.
EWC~\cite{kirkpatrick2017overcoming} is a classic regularization-based method by restricting parameter changes across tasks in the local region.
PASS~\cite{zhu2021prototype} and SSRE~\cite{zhu2022self} aim to train a more balanced classifier with the pseudo features sampled based on the old class statistics.
A distribution-based classifier is exploited by IL2A~\cite{zhu2021class} for the classifier learning of the old classes, while the new classifier is separately trained with the given samples.
Furthermore, the feature extractors in FeTrIL~\cite{petit2023fetril} and FeCAM~\cite{goswami2023fecam} are trained at the first task only and fixed at the following incremental tasks, which are different from the optimization paradigms of other methods.
A parameterized classifier head, i.e., a fully connected (FC) layer, is learned by FeTrIL~\cite{petit2023fetril}, while FeCAM~\cite{goswami2023fecam} classifies with a distance metric based on covariance matrices.
The results of competitors are reproduced.

\begin{table*}[t]
\caption{
Overall performance 
of different models.
The best results are in red, and the second best in blue.
}
\label{table:main_result}
\vskip 0.15in
\begin{center}
\begin{small}
\begin{sc}
\begin{tabular}{l |c cccc| c cccc| c cccc}
\hline
\multirow{3}{*}{Method} & &\multicolumn{4}{c|}{ImageNet-S} & & \multicolumn{4}{c|}{Tiny-ImageNet} & & \multicolumn{4}{c}{CIFAR100}\\
& & \multicolumn{2}{c}{T=10} & \multicolumn{2}{c|}{T=20} &  & \multicolumn{2}{c}{T=10} & \multicolumn{2}{c|}{T=20} &  & \multicolumn{2}{c}{T=10} & \multicolumn{2}{c}{T=20}\\
& & LA & \multicolumn{1}{c}{AIA} & LA & AIA & & LA & \multicolumn{1}{c}{AIA} & LA & AIA & & LA & \multicolumn{1}{c}{AIA} & LA & AIA\\
\hline
EWC & &  10.4 & 28.6 & 5.5 & 19.1 &  & 9.7 & 25.0 & 4.9 & 16.6 &  & 9.7 & 27.2 & 5.5 & 19.5\\
IL2A & & 34.2 & 46.2 & 17.2 & 28.7 &  & 3.0 & 7.9 & 2.2 & 7.6 &  & 30.4 & 39.9 & 6.0 & 14.4\\
PASS & & 28.2 & 43.8 & 12.9 & 23.9 &   & 4.7 & 10.0 & 0.5 & 1.8 & & 34.9 & 49.0 & 19.0 & 28.7\\
SSRE & & 25.8 & 42.1 & 21.3 & 36.8 &  & 27.1 & 27.1 & 13.1 & 23.1 & & 32.1 & 45.9 & 17.7& 31.8 \\
FeTrIL & & 28.6 & 47.2 & 21.3 & 39.0 &  & 24.1 & 38.0 & 17.1 & 29.7 &  & \textcolor{blue}{32.7} & \textcolor{blue}{49.4} & \textcolor{blue}{26.1} &\textcolor{blue}{42.2}\\
FeCAM & & \textcolor{blue}{36.2} & \textcolor{blue}{53.5} & \textcolor{blue}{31.2} &\textcolor{blue}{45.6} &  & \textcolor{blue}{27.6} & \textcolor{blue}{40.6} & \textcolor{blue}{17.7} &\textcolor{blue}{30.1} &  & 31.7 & 48.0 & 25.0 & 41.1 \\
AMGC & & \textcolor{red}{39.9} & \textcolor{red}{55.1} & \textcolor{red}{32.9} & \textcolor{red}{47.2} &  & \textcolor{red}{28.3} & \textcolor{red}{41.3} & \textcolor{red}{18.2} &\textcolor{red}{30.7} &  & \textcolor{red}{36.2} & \textcolor{red}{51.7} & \textcolor{red}{30.8} &\textcolor{red}{42.5}\\
\hline
\end{tabular}
\end{sc}
\end{small}
\end{center}
\vskip -0.1in
\end{table*}

\subsection{Main Results}
The results in Table~\ref{table:main_result} demonstrate the effectiveness of the proposed AMGC by its state-of-the-art (SOTA) level performance across various settings.
AMGC outperforms its counterparts, including EWC, IL2A, PASS, and SSRE, which continually update feature extractors and classifier heads during the incremental learning process. For instance, when compared to SSRE, the performance of AMGC under \cxb{20-task} \yzr{ImageNet-S} is \cxb{more than $10\%$ better on both criteria.}

In contrast, FeTrIL and FeCAM are methods that only train classifier heads at incremental tasks while keeping their feature extractor frozen at initial states. These approaches achieve better results than the holistic updating models mentioned above, except for AMGC.
This phenomenon reflects the challenge of continually learning a feature extractor under EFCIL. Such an incremental learning process is vulnerable to catastrophic forgetting, characterized by classifier biases and deteriorated old class features.
The proposed AMGC is neat and effective \cxb{in handling these challenges.
AMGC is consistently better than FeTrIL and FeCAM and achieves state-of-the-art results, as shown in Table~\ref{table:main_result}.}
For example, AMGC outperforms FeCAM by $1.6\%$ \cxb{AIA}
on both T = 10 and T = 20 settings of the \yzr{ImageNet-S}.
\cxb{Corresponding improvements in LA enlarge to $3.7\%$ and $1.7\%$, respectively.
}



\subsection{Detailed Analysis}

\textbf{Ablation Study.}
Our AMGC consists of two parts: 
\yzri{DBGC and AMarX.}
AMarX is built upon DBGC.
DBGC has four main variants: SBLC, SB$^{n}$DB$^{o}$LC, SBGC, and DBLC. \footnote{\yzri{The definition of each variant can be found in supplementary material section B.}}
The 
\yzri{SBLC}
is neither DB nor GC and thus serves the fully ablative variant of DBGC.
As shown in Tabel~\ref{table:ablation}, SBLC obtains the worst results, as it suffers from sampling bias and local optima.
SB$^{n}$DB$^{o}$LC is introduced by IL2A with the classifier of old classes belonging to DB. 
Using DBLC, classifiers for old and new classes are separately learned based on the DB loss to defy sampling bias, resulting in substantial improvements.
The proposed DBGC further improves DBLC by learning a holistic classifier for both old and new classes. DBGC is $6.0\%$ and $16.7\%$ higher \yzr{in AIA} than DBLC on the \yzr{ImageNet-S} and CIFAR100, respectively.
\cxb{Larger improvements in LA can also be observed.}
We combine AMarX with DBGC to get the complete model AMGC. AMGC achieves the best results and further boosts DBGC by at least \cxb{$1.1\%$ LA and} $1.4\%$ \cxb{AIA}.

\begin{table}[t]
\caption{
Ablation of the performance indicates the contributions from different components of the proposed AMGC.
\yzr{ImageNet-S} and CIFAR100 T = 20 settings are used. 
\cxb{The best results are in red, and the second best in blue.}
}

\label{table:ablation}
\vskip 0.15in
\begin{center}
\begin{small}
\begin{sc}
\begin{tabular}{l c|cc| cc}
\hline
\multirow{2}{*}{Method} &  & \multicolumn{2}{c|}{ImageNet-S}  & \multicolumn{2}{c}{CIFAR100}\\
 & & LA & AIA & LA & AIA \\
\hline
SBLC &  & 5.1  & 17.0  & 4.8 & 17.3  \\

SB$^{n}$DB$^{o}$LC & & 5.0 &   20.3  & 5.6 &  18.9 \\
SBGC &  & 8.3 & 24.4  & 4.8 & 17.3  \\
DBLC &  & 25.0 & 39.4  & 8.3 & 24.4  \\

DBGC &  & \textcolor{blue}{32.9}  & \textcolor{blue}{45.4}  & \textcolor{blue}{29.2} & \textcolor{blue}{41.1} \\
AMGC &  & \textcolor{red}{33.0} & \textcolor{red}{47.2}   & \textcolor{red}{30.6} & \textcolor{red}{42.5} \\
\hline
\end{tabular}
\end{sc}
\end{small}
\end{center}
\vskip -0.1in
\end{table}

\textbf{Different Types of Margins.}
The proposed AMarX is proven to be a cross-entropy with an adaptive class-specific margin $m_k$ for the class $k$.
Existing classification loss with margin is usually defined with a class agnostic hyper-parameter $m$.
We choose the frequently-used 
Soft-Margin (SM)~\cite{liang2017soft} as an alternative to AMarX and apply it on DBGC for the old classes.
\cxb{However, DBGC with SM fails to bring any improvement and even harms the performance, as shown in Table~\ref{table:dif-margin}.}
Additional experiments and analyses are available in the supplementary \yzr{section D}. 

\begin{table}[t]
\caption{
Losses with different margins based on our DBGC.
SM refers to Soft-Margin. 
\yzr{ImageNet-S} and CIFAR100 T=20 settings are used. 
}\label{table:dif-margin}
\vskip 0.15in
\begin{center}
\begin{tabular}{lccccc}
\hline
\multirow{2}{*}{Margin type} & &  \multicolumn{2}{c}{ImageNet-S} & \multicolumn{2}{c}{CIFAR100}\\
&  & LA & AIA & LA &  AIA\\
\hline
 DBGC     & &  32.9 & 45.4   & 29.2 & 41.1     \\
\yzr{DBGC}+SM   & &  24.2 & 40.7   & 18.6 &35.6    \\
 AMGC & &  33.0 &  47.2     & 30.6 & 42.5   \\
\hline
\end{tabular}
\end{center}
\vskip -0.1in
\end{table}

\section{Conclusion}
In this paper, the proposed method targets two challenges 
in 
\yzri{EFCIL}
, resulting in the following main contributions.
Firstly, 
\yzri{DBGC}
is introduced to alleviate the learning biases found in existing EFCIL methods.
Secondly, the proposed method considers the degradation of old class features under EFCIL and simulates it via 
\yzri{VE.}
We show that applying VE along with DBGC is equivalent to introducing the class-specific margins into the classification loss, resulting in 
\yzri{AMarX.}
Our full model comprises DBGC and AMarX, called 
\yzri{AMGC}.
Extensive experiments under a challenging EFCIL setting are conducted to demonstrate the superiority of AMGC.

\yzri{
\textbf{Acknowledgement} This research is supported by the National Science Foundation for Young Scientists of China (No. 62106289).
}

\end{document}